\documentclass[10pt,twocolumn,letterpaper]{article}
\PassOptionsToPackage{table}{xcolor}
\usepackage[pagenumbers]{cvpr}
\usepackage{multirow}
\usepackage{bm}
\usepackage{microtype}
\definecolor{cvprblue}{rgb}{0.21,0.49,0.74}
\usepackage[breaklinks,colorlinks,allcolors=cvprblue]{hyperref}
\author{\parbox{0.97\textwidth}{\centering
\vspace{1em}
{\bfseries
Yatai~Ji\textsuperscript{\rm 1, \rm 2}, Zhengqiu~Zhu\textsuperscript{\rm 1, \rm 2}\thanks{Corresponding author.}, Yong~Zhao\textsuperscript{\rm 1, \rm 2}, Yue~Hu\textsuperscript{\rm 1, \rm 2}, Fanglong~Yao\textsuperscript{\rm 3}, Chen~Gao\textsuperscript{\rm 4}, Pengfei~Zhu\textsuperscript{\rm 5}, Quanjun~Yin\textsuperscript{\rm 1, \rm 2}\\}
\vspace{1em}
\normalfont
\textsuperscript{\rm 1}National Key Laboratory of Digital Intelligent Modeling and Simulation, National University of Defense Technology\\
\textsuperscript{\rm 2}College of Systems Engineering, National University of Defense Technology\\
\textsuperscript{\rm 3}Aerospace Information Research Institute, Chinese Academy of Sciences\\
\textsuperscript{\rm 4}BNRist, Tsinghua University\\
\textsuperscript{\rm 5}Southeast University\\
\vspace{1em}
\{zhuzhengqiu12@nudt.edu.cn\}}}
\hypersetup{pdfauthor={Yatai Ji, Zhengqiu Zhu, Yong Zhao, Yue Hu, Fanglong Yao, Chen Gao, Pengfei Zhu, Quanjun Yin}}

\hypersetup{pdftitle={SearchWorld: Spatial Value-Guided Imagination for UAV Object Search via World Models},pdfsubject={},pdfkeywords={}}
\title{SearchWorld: Spatial Value-Guided Imagination for UAV Object Search via World Models}
\newcommand{\method}{{\upshape\textsc{SearchWorld}}}
\newcommand{\Mexpl}{\mathbf{M}^{\mathrm{expl}}}

\AddToHook{cmd/@maketitle/after}{\begin{minipage}{\textwidth}
\centering
\includegraphics[width=0.92\linewidth]{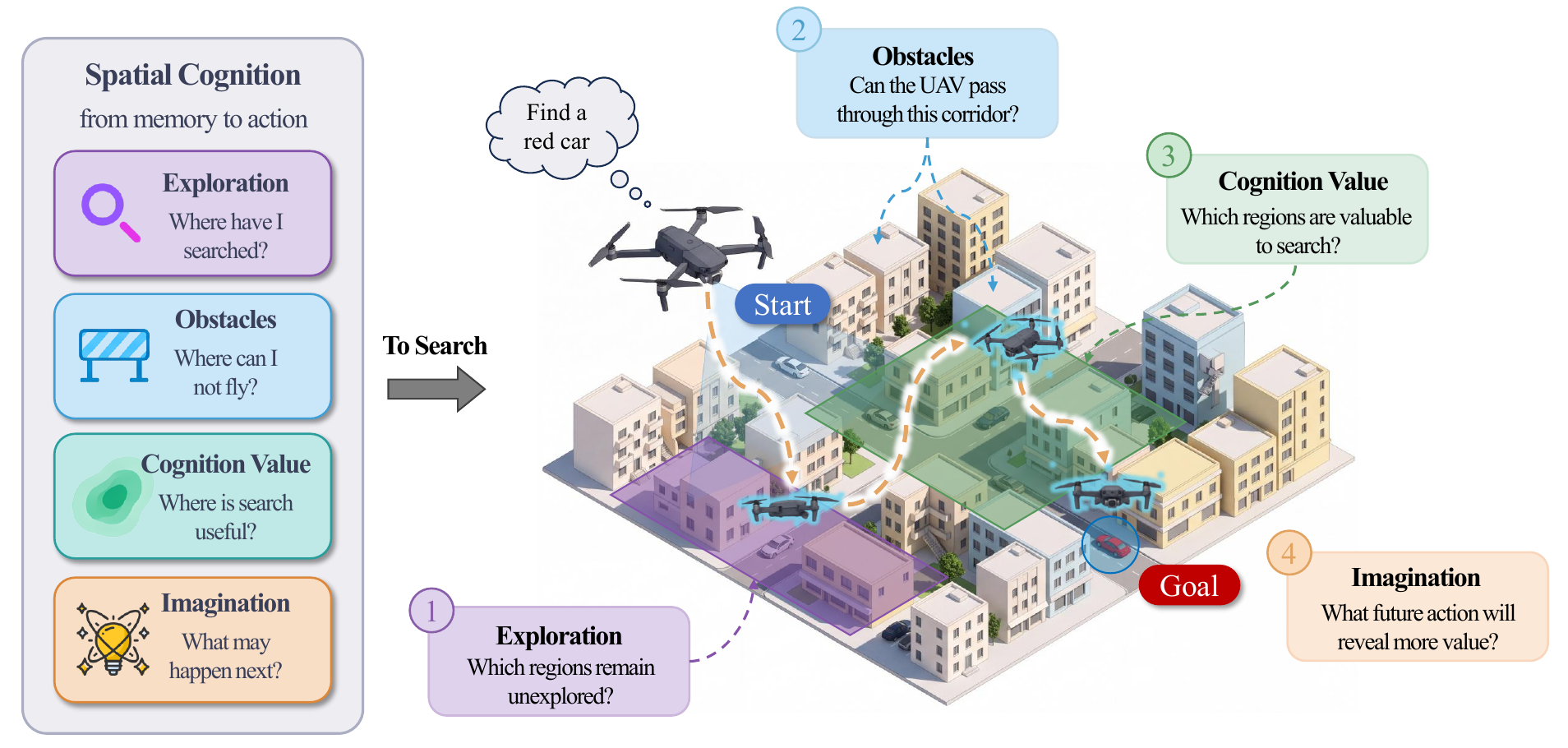}
\captionsetup{type=figure}
\caption{Spatial memory guides training-time imagination in \method{}.}
\label{fig:teaser}
\end{minipage}
\par\vspace{12pt}
}
\begin{document}
\maketitle
\begin{abstract}
Urban unmanned aerial vehicle (UAV) object search requires finding an instruction-specified target from partial observations. Spatial memory records where the UAV has searched, but learning how future actions affect search utility remains challenging. We propose SearchWorld, a recurrent state-space world model that connects explicit spatial memory with value-guided imagination. It maintains bird's-eye-view exploration and obstacle layers and decodes a task-aware spatial value layer from latent states. A cognition--action network is designed to turn these spatial utilities into policy improvements through imagined rollouts during training, without a separate scalar critic. The three-stage curriculum first learns dynamics and spatial representations, then establishes target-seeking and stopping behavior through expert imitation, and finally refines movement actions using imagined utilities. This sequence provides imagination-based refinement with both a learned world model and a task-specific behavioral prior. At test time, the refined policy directly selects actions without imagined rollouts. On UAV-ON, \method{} achieves $23.8\%$ success ($19.5\%$ for OctMem-Agent) and $35.5\%$ oracle success, with $23.0\%$ success on unseen scenes. These results demonstrate the effectiveness of connecting explicit spatial memory with training-time imagination for aerial object search.

\end{abstract}
\section{Introduction}
\label{sec:intro}
Autonomous unmanned aerial vehicles (UAVs) are increasingly used to search for objects in large urban environments. In aerial object search, the UAV often receives an RGB-D view, its pose, and an instruction describing a target instance, then must find and stop at that target without an externally provided route. Unlike conventional indoor ObjectNav and short-horizon aerial navigation, outdoor search provides no target location and may contain many similar objects across a region much larger than the camera field of view. The agent must maintain what has been observed while deciding which future region is most useful to visit. Broad exploration wastes the step budget while following only the current target hypothesis can cause repeated visits and occlusion traps.

Map-based agents mitigate partial observability with explicit spatial memory \citep{yamauchi1997frontier, chaplot2018active, chaplot2020semexp}. Spatial maps and scene representations \citep{huang2023vlmaps, gu2024conceptgraphs} organize accumulated observations, but do not by themselves specify how those observations evolve under candidate actions. World models provide this predictive capability through learned transitions \citep{ha2018worldmodels}. For aerial search, the challenge is to connect these transitions to the spatial questions that guide exploration: \textbf{which regions have been observed, which are obstructed, and where the instruction-specified target may be found}. A scalar return estimate summarizes utility, whereas a spatial utility field exposes its distribution over locations and allows action-dependent geometric readout. Our design couples this readout with latent imagination to construct training targets for the movement policy.

We propose \textbf{SearchWorld}, a recurrent state-space world model grounded in explicit spatial cognition memory (Figure~\ref{fig:teaser}). Specifically, we organize this memory as a three-layer bird’s-eye-view (BEV) representation: an exploration layer records observed coverage, an obstacle layer marks projected obstacles, and a spatial value layer decoded from the latent state estimates where searching is worthwhile. We further introduce a \textbf{cognition–action network}, in which the decoded spatial value guides training-time imagination to improve the agent’s action policy. A three-stage curriculum trains SearchWorld to learn state transitions and spatial representations, imitate expert search behavior, and refine spatial exploration through value-guided imagination. At test time, the refined policy directly selects actions from observation-grounded latent states. On UAV-ON, SearchWorld achieves 23.8\% SR and 35.5\% OSR overall, exceeds the listed baselines in SR and OSR, and retains 23.0\% success on unseen scenes. Our contributions are:

\begin{itemize}
\item \textbf{We design a spatially grounded world model for aerial search.} SearchWorld combines a recurrent state-space model (RSSM) with allocentric exploration and obstacle channels and decodes a task-aware spatial value layer from the latent state.
\item \textbf{We propose a cognition--action network for value-guided imagination.} A decoded task-aware spatial value layer, supervised by exploration, obstacle, and training-time target signals but not given the target coordinate at inference, replaces the actor--critic's learned scalar critic as a structured value prior. During training, imagination-based value guidance refines movement actions by pooling decoded value over imagined footprints, with the imitated stopping behavior kept intact.
\item \textbf{We develop a three-stage training curriculum and evaluate it on UAV-ON.} A three-stage curriculum separates world-model learning, expert task imitation, and value-guided movement refinement. Experiments on UAV-ON evaluate overall and held-out-scene search performance, with ablations examining architecture, spatial memory, and training stages.
\end{itemize}

\section{Related Work}
\label{sec:related}
\paragraph{World models for navigation.} One line of world models plans navigation by generating future observations as pixels. PathDreamer \citep{koh2021pathdreamer} synthesizes novel indoor viewpoints, DreamWalker \citep{wang2023dreamwalker} imagines candidate plans with a learned scene synthesizer, DreamNav \citep{wang2025dreamnav} and NWM \citep{bar2025nwm} simulate candidate trajectories with large generative models, and ANWM \citep{zhang2025anwm} extends such video prediction to aerial trajectories. A second line uses latent world models, with DreamerV3 \citep{hafner2023dreamerv3} learning behavior through imagined latent trajectories. MILE \citep{chen2022mile} learns driving dynamics in latent space, while X-Mobility \citep{liu2024xmobility} and NavMorph \citep{yao2025navmorph} carry RSSMs into outdoor and vision-and-language navigation. WMNav \citep{nie2025wmnav} instead uses a vision--language model (VLM) to predict action outcomes and update an explicit online curiosity value map. \method{} connects an action-conditioned RSSM to sensor-built spatial memory and a decoded utility layer, using imagined utilities to train a policy that executes directly.

\paragraph{Aerial navigation and object search.} Aerial navigation research has progressed from instruction-following flight (AerialVLN~\citep{liu2023aerialvln}; CityNav~\citep{lee2024citynav}) to object-goal search, formalized by UAV-ON \citep{xiao2025uavon}. CityEQA \citep{zhao2025cityeqa} studies question answering through active urban exploration, while CityCube \citep{xu2026citycube} evaluates cross-view spatial reasoning in urban environments. VLFM \citep{yokoyama2024vlfm} uses language-grounded value maps to select navigation frontiers. Recent aerial search agents couple explicit spatial memory with large language models (LLMs) and their multimodal variants (MLLMs). APEX \citep{zhang2026apex} combines attraction, exploration, and obstacle maps with a policy trained by proximal policy optimization (PPO) and asynchronous target grounding; PRPSearcher \citep{ji2026cityavos} has an MLLM reason over semantic-attraction and uncertainty maps; and OctMem-Agent \citep{zhou2026octmem} aggregates the RGB-D history into an adaptive octree memory queried by instruction-modulated tokens. These approaches demonstrate the usefulness of spatial memory for aerial search. \method{} focuses on coupling such memory with learned latent transitions and training-time spatial-utility distillation.

\section{Method}
\label{sec:method}

\begin{figure*}[t]
    \centering
    \includegraphics[
        width=0.98\linewidth
    ]{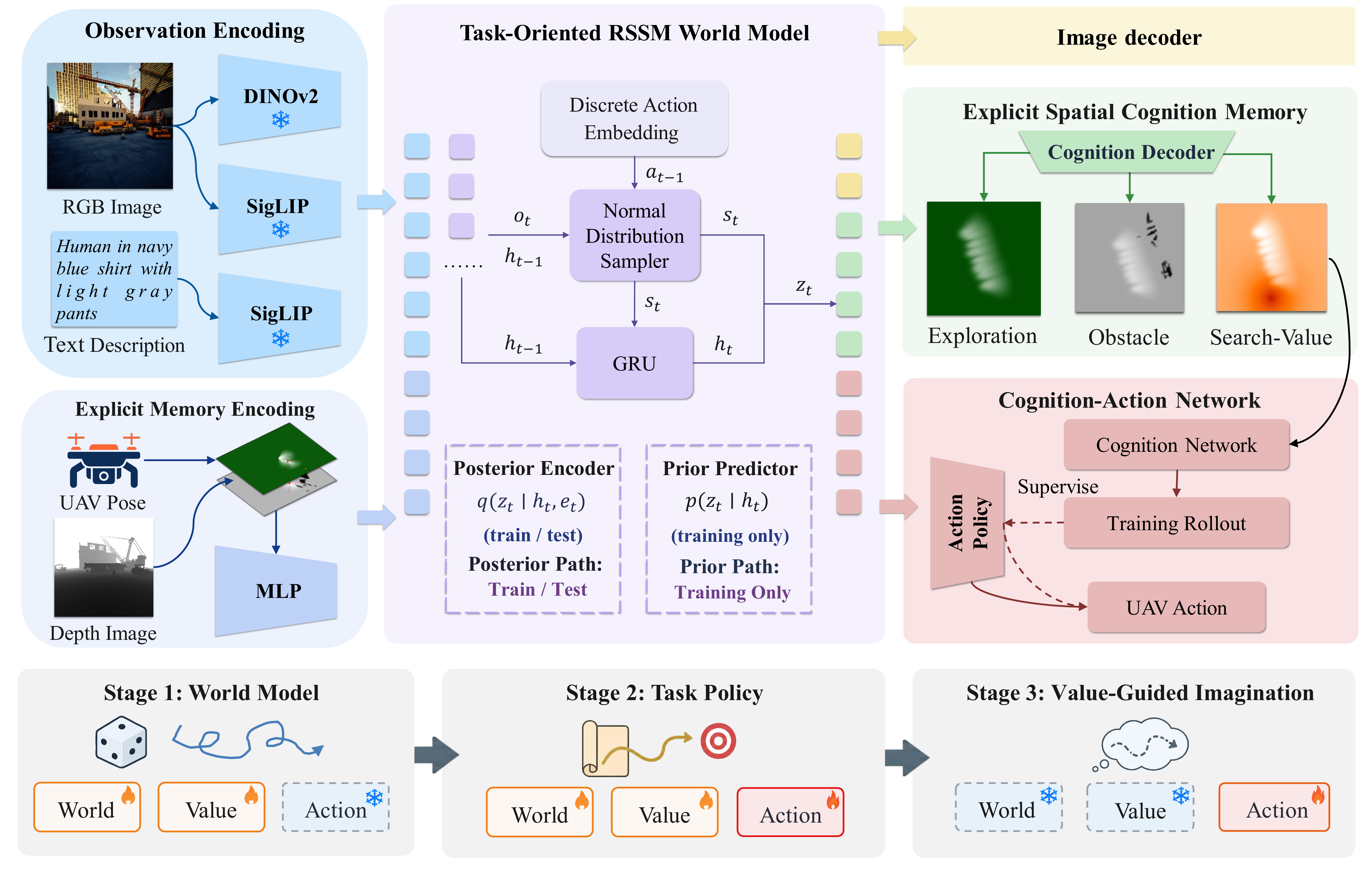}
    \caption{\method{} overview with its three-stage curriculum. A task-oriented RSSM world model encodes the RGB view, the instruction, and the explicit spatial-cognition memory, and decodes the observation, the spatial memory, the task-aware value layer, and the UAV action. The cognition--action network trains the action policy in Stage~2 and refines movement actions in Stage~3 by pooling the decoded value over imagined action footprints. Rollout-based guidance is used during training; at test time, the posterior incorporates real observations and the refined action policy directly outputs actions.}
    \label{fig:pipeline}
\end{figure*}

\method{} rests on three components. An \emph{explicit spatial-cognition memory} makes the search state inspectable: exploration and obstacle layers are constructed from depth and pose, and a value layer is decoded from the world model's latent state (Section~\ref{sec:bev_memory}). A \emph{task-oriented RSSM world model} learns the search dynamics over this memory, predicts future states in training-time imagination, and decodes their value layers (Section~\ref{sec:world_model}). A \emph{cognition--action network} distills the decoded spatial value into an action policy during training: the task policy learned from experts is refined by value-guided updates that touch movement actions only (Section~\ref{sec:value_imagination}). Section~\ref{sec:formulation} formalizes the problem, and Section~\ref{sec:training} presents the three-stage curriculum that trains the components. Figure~\ref{fig:pipeline} gives an overview.

\subsection{Problem Formulation}
\label{sec:formulation}

We model aerial object-goal search as a partially observable Markov decision process (POMDP). The UAV's own state, including its 6-DoF pose $p_t$, is known at each step. Partial observability concerns the environment layout and target location. The observation $o_t = (I_t, D_t, p_t, \ell)$ comprises RGB and depth images, the known pose, and an instance-level text instruction $\ell$ describing the target. Based on these observations, the UAV executes a learned policy to locate the instruction-specified target and stop in its vicinity. The action space consists of eight discrete actions: \texttt{forward}, \texttt{left}, \texttt{right}, \texttt{ascend}, \texttt{descend}, \texttt{rotate-left}, \texttt{rotate-right}, and \texttt{stop}. 

\subsection{Explicit Spatial-Cognition Memory}
\label{sec:bev_memory}

A recurrent latent state can summarize interaction history, but does not expose a physically indexed record of spatial coverage. \method{} therefore maintains an \emph{explicit spatial-cognition memory} to represent observed regions and projected obstacles. The memory uses a $256\times256$ BEV grid with $0.4\,\mathrm{m}$ cells, spanning $102.4\times102.4\,\mathrm{m}$ and centered on the search region. World points map to cells by a fixed affine rule, keeping the allocentric coordinate system consistent across steps. Additional geometric parameters are given in Table~\ref{tab:specs} in Appendix~\ref{app:implementation}.

The memory has three layers with distinct origins. Fixed geometric rules construct the exploration and obstacle layers from pose and depth; their resolution and sensing geometry determine the spatial information retained. The value layer is an \emph{output of the world model}, decoded from the latent state and conditioned on the search instruction. It scores cells by the predicted utility of visiting them under the current task, using task-aware supervision available during training (Section~\ref{sec:value_imagination}). The first two layers summarize observations, while the third estimates where searching is worthwhile.
\paragraph{Exploration layer: where have I observed?} Each executed step projects a field of view (FOV) of angular width $\phi$ and sensing range $\rho$. For a cell $c$, let $d_\tau(c)$ and $\Delta\theta_\tau(c)$ denote its range and bearing offset relative to the pose $p_\tau$ at step $\tau$. The layer accumulates weighted FOV coverage up to step $t$,
\begin{equation}
\label{eq:expl}
\begin{aligned}
\Mexpl(c) ={}& \max_{\tau \le t}\Big\{\Big(1 - \frac{d_\tau(c)}{\rho}\Big)
\cos\Big(\frac{\pi \Delta\theta_\tau(c)}{\phi}\Big) \\
&{}\cdot \mathbb{1}\big[d_\tau(c) \le \rho,\; |\Delta\theta_\tau(c)| \le \phi/2\big]\Big\},
\end{aligned}
\end{equation}
where the temporal maximum retains coverage and the linear/cosine weights reduce the contribution of distant or oblique regions.

\paragraph{Obstacle layer: where are obstacles observed?} The front-view depth stream $D_t$ is back-projected to a 3-D point cloud with the known camera intrinsics, transformed to world coordinates by the pose $p_t$, and orthographically projected onto the grid. A cell is marked obstacle if the projected points within it exceed a height threshold $z_{\mathrm{obs}}$ above the ground. The resulting mask supplies a compact geometric constraint for search.

\paragraph{Properties.} Both observation-level layers are \emph{explicit}: every cell has a physical coordinate and an interpretable meaning. They are also \emph{complementary}: exploration accumulates coverage, while the obstacle mask records projected geometry. Their construction requires no learned parameters; the resulting maps serve as both observation inputs and reconstruction targets for the world model.

\subsection{Task-Oriented RSSM World Model}
\label{sec:world_model}

The core of \method{} is a task-oriented world model of the search process: a compact recurrent dynamics model that rolls the interaction forward \emph{in imagination during training}, without querying the simulator. Figure~\ref{fig:pipeline} shows its structure, with every component aligned to the structure of object search.

\paragraph{Observation encoding.} The world model consumes three input streams. The RGB image $I_t$ is encoded by two frozen encoders: a DINOv2 vision transformer (ViT) \citep{oquab2023dinov2} extracts spatial and geometric structure, and the image tower of a SigLIP \citep{zhai2023siglip} extracts language-aligned features. The frozen encoders provide reusable representations, an instance of feature-based model reuse \citep{li2025deep}. The text instruction $\ell$ is encoded by the \emph{text tower of the same SigLIP}. Sharing one encoder family between image and text places what the camera sees and what the instruction asks for in a common embedding space, so the world model can learn which visual evidence is relevant to the target. The exploration and obstacle layers of the spatial-cognition memory (Section~\ref{sec:bev_memory}) enter as the third stream through a BEV encoder. All streams are fused into the observation embedding $e_t$.

\paragraph{Recurrent state-space dynamics.} Following the RSSM family \citep{hafner2019planning, hafner2020dreamer}, the world model maintains a deterministic recurrent state $h_t$ and a stochastic latent $z_t$, factorizing the belief into what is predictable and what remains uncertain. The deterministic path is updated by a gated recurrent unit (GRU),
\begin{equation}
\label{eq:gru}
h_t = \mathrm{GRU}\big(z_{t-1},\, \mathrm{Emb}(a_{t-1}),\, h_{t-1}\big),
\end{equation}
where the discrete action is embedded by a learned lookup table, a simple interface that matches the atomic action space of Section~\ref{sec:formulation}. Two heads predict the stochastic state. The \emph{posterior} (state estimator) $q_\theta(z_t \mid h_t, e_t)$ conditions on the current observation; the \emph{prior} (state predictor) $p_\theta(z_t \mid h_t)$ receives none. The posterior incorporates real observations during both training and deployment. The prior predicts observation-free successor states for training-time imagination.

\paragraph{Decoders.} Decoder heads read the task signals out of the latent state. An RGB decoder regenerates the observation. A BEV decoder regenerates the exploration and obstacle layers of the spatial-cognition memory, and a value decoder regenerates the task-aware value layer under the supervision target of \eqref{eq:valuetarget}. An action head outputs a categorical distribution over the eight atomic actions, conditioned on the latent state and the instruction (Section~\ref{sec:training}). During Stage~3, imagined rollouts provide spatial utilities for policy refinement; the observation and value decoders are not required for test-time action selection.

\paragraph{Training-time imagination via the prior path.} We define the \emph{imagination step} as one prior-only transition. Given an observation-grounded belief $(h_t, z_t)$ and a candidate action $a_t$ during training,
\begin{equation}
\label{eq:imagine}
h_{t+1} = \mathrm{GRU}(z_t, \mathrm{Emb}(a_t), h_t), \qquad
z_{t+1} \sim p_\theta(\cdot \mid h_{t+1}).
\end{equation}
Iterating \eqref{eq:imagine} rolls out hypothetical trajectories purely in latent space, scored at each step by the decoder heads. These rollouts are used for the Stage-3 policy learning of Section~\ref{sec:training}.

\subsection{Cognition--Action Network}
\label{sec:value_imagination}

Dreamer-style model-based RL \citep{hafner2020dreamer} learns an independent scalar critic $V_\psi(s_t)$ by regressing imagined returns. Under rewards as sparse as aerial search success, return-based value learning can be difficult to stabilize. \method{} instead replaces the actor--critic with a \emph{cognition--action network}. Its cognition branch decodes a structured spatial value prior from the world model's latent state. During training, the resulting imagined utilities supervise the instruction-conditioned action branch.

\paragraph{Cognition branch: the decoded task-aware spatial value layer.} Given the RSSM state $s_t = (h_t, z_t)$ and the instruction encoding $e_\ell$, the cognition decoder $D_V$ predicts a spatial value layer over the allocentric BEV grid,
\begin{equation}
\label{eq:value}
\hat V_t^{\mathrm{sp}} \;=\; D_V(h_t, z_t, e_\ell),
\end{equation}
where $\hat V_t^{\mathrm{sp}}(x)$ is the predicted utility of visiting location $x$, realized as the belief field of the cognitive map. The value layer is decoded rather than supplied as an input; its supervision target is constructed only during training. With $M_t^{\mathrm{expl},*}$ and $M_t^{\mathrm{obs},*}$ the geometrically constructed exploration and obstacle layers of Section~\ref{sec:bev_memory} (the training-time ground-truth labels, starred to distinguish them from the decoder's outputs), and $g$ the ground-truth target position available during training, the supervision target gives explored cells lower value, obstacle cells zero, and cells near the target higher value through a Gaussian prior,
\begin{equation}
\label{eq:valuetarget}
\begin{aligned}
V_t^{*}(x) = \operatorname{Norm}\Big[&\big(1 - M_t^{\mathrm{obs},*}(x)\big) \\
{}\cdot\Big(&\lambda_{\mathrm{expl}}\big(1 - M_t^{\mathrm{expl},*}(x)\big)
+ \lambda_{\mathrm{goal}}G_g(x)\Big)\Big],
\end{aligned}
\end{equation}
Here $G_g(x) = \exp\big(-\|x-g\|_2^2/2\sigma_g^2\big)$, with $\lambda_{\mathrm{expl}}=\lambda_{\mathrm{goal}}=1.0$ and $\sigma_g=10\,\mathrm{m}$; $\operatorname{Norm}$ sets unit maximum. We minimize
\begin{equation}
\label{eq:valueloss}
\mathcal{L}_{\mathrm{value}} \;=\; \frac{1}{|\mathcal{G}|} \sum_{x \in \mathcal{G}} \ell_{\mathrm{rob}}\big(\hat V_t^{\mathrm{sp}}(x) - V_t^{*}(x)\big),
\end{equation}
where $\mathcal{G}$ is the BEV grid and $\ell_{\mathrm{rob}}$ a robust regression loss. The target coordinate $g$ is used to construct training labels and is never an input to the observation encoder, value decoder, or deployed policy. During Stage~3, decoding \emph{imagined} successor states provides utilities for candidate movement actions.

\paragraph{Action branch.} The action branch maps the latent state and instruction to a categorical distribution over the eight atomic actions. It is initialized by the Stage-2 behavior-cloned policy $\pi_{\mathrm{BC}}$ and refined by the value-guided objective of Section~\ref{sec:training}. The update applies to the movement actions $\mathcal{A}_{\mathrm{move}}$, preserving $\pi_{\mathrm{BC}}(\texttt{stop}\mid s_t,e_\ell)$ at a given belief. This retains the imitated stopping rule while refining movement; different movements can still lead to different subsequent observations and stopping locations.

\begin{table*}[!t]
\centering
\caption{Comparison on UAV-ON by target size and overall. SR, OSR, and SPL are reported in \%; higher is better. Bold and underlining mark the best and second-best results in each column, respectively; -- denotes an unreported size breakdown. Published baselines retain their original control settings (Appendix~\ref{app:metrics}).}
\label{tab:object_size_performance}

\small
\setlength{\tabcolsep}{3pt}
\begin{tabular}{lcccccccccccc}
\toprule
\multirow{2}{*}{\textbf{Method}}
& \multicolumn{3}{c}{\textbf{Small}}
& \multicolumn{3}{c}{\textbf{Medium}}
& \multicolumn{3}{c}{\textbf{Large}}
& \multicolumn{3}{c}{\textbf{Total}} \\
\cmidrule(lr){2-4}
\cmidrule(lr){5-7}
\cmidrule(lr){8-10}
\cmidrule(lr){11-13}

& SR$\uparrow$ & OSR$\uparrow$ & SPL$\uparrow$
& SR$\uparrow$ & OSR$\uparrow$ & SPL$\uparrow$
& SR$\uparrow$ & OSR$\uparrow$ & SPL$\uparrow$
& SR$\uparrow$ & OSR$\uparrow$ & SPL$\uparrow$ \\
\midrule

Random
& 4.14 & 7.80 & 2.80
& 3.33 & 8.10 & 3.05
& 2.48 & 8.07 & 1.62
& 3.70 & 8.00 & 2.66 \\

CLIP-H
& 2.86 & 8.43 & 1.51
& 10.95 & 16.67 & 7.17
& 13.04 & 19.25 & 10.53
& 6.20 & 11.90 & 4.15 \\

AOA-V
& 2.86 & 25.44 & 0.54
& 5.71 & \underline{27.62} & 1.73
& 7.45 & 27.95 & 1.04
& 4.20 & 26.30 & 0.87 \\

AOA-F
& 4.45 & 16.38 & 1.61
& 10.48 & 17.62 & 6.36
& 14.29 & 21.74 & \underline{10.66}
& 7.30 & 17.50 & 4.06 \\

WMNav
& 5.55 & 20.87 & 1.77
& 9.04 & 23.91 & 3.30
& 16.20 & 28.29 & 6.56
& 8.00 & 22.70 & 2.86 \\

APEX
& -- & -- & --
& -- & -- & --
& -- & -- & --
& 13.33 & 20.00 & \textbf{10.14} \\

PRPSearcher
& 8.67 & 24.71 & 4.21
& 13.03 & 23.71 & \underline{7.34}
& 17.50 & \underline{32.15} & 8.82
& 11.00 & 25.70 & 5.61 \\

OctMem-Agent
& \underline{18.91} & \underline{29.57} & \underline{6.03}
& \underline{18.09} & 26.66 & 6.27
& \underline{23.60} & 31.67 & 7.80
& \underline{19.50} & \underline{29.30} & 6.37 \\
\midrule
\rowcolor{blue!8}
\textbf{\method{} (Ours)}
& \textbf{21.60} & \textbf{35.25} & \textbf{7.23}
& \textbf{21.80} & \textbf{34.64} & \textbf{8.35}
& \textbf{35.02} & \textbf{37.59} & \textbf{10.76}
& \textbf{23.80} & \textbf{35.50} & \underline{8.03} \\

\bottomrule
\end{tabular}
\end{table*}

\begin{table*}[!t]
\begin{minipage}[t]{0.52\textwidth}
    \centering
    \caption{Scene generalization on UAV-ON (\%).}
    \label{tab:seen_unseen_comparison}
    \setlength{\tabcolsep}{3.2pt}
    \renewcommand{\arraystretch}{1.05}
    \resizebox{\linewidth}{!}{%
        \begin{tabular}{lccc@{\hspace{8pt}}ccc}
            \toprule
            \textbf{Method}
            & \multicolumn{3}{c}{\textbf{Seen}}
            & \multicolumn{3}{c}{\textbf{Unseen}} \\
            \cmidrule(lr){2-4}
            \cmidrule(lr){5-7}
            & \textbf{SR$\uparrow$}
            & \textbf{OSR$\uparrow$}
            & \textbf{SPL$\uparrow$}
            & \textbf{SR$\uparrow$}
            & \textbf{OSR$\uparrow$}
            & \textbf{SPL$\uparrow$} \\
            \midrule
            WMNav
            & 8.69 & 23.60 & 3.20
            & 7.59 & 22.16 & 2.66 \\
            APEX
            & 12.36 & 19.10 & \textbf{9.13}
            & 16.13 & 22.58 & \textbf{13.03} \\
            PRPSearcher
            & 10.71 & 23.46 & 6.26
            & 11.17 & 27.04 & 5.22 \\
            OctMem-Agent
            & \underline{22.72} & \underline{30.48} & \underline{8.40}
            & \underline{17.57} & \underline{28.59} & 5.15 \\
            \midrule
            \rowcolor{blue!8}
            \textbf{\method{} (Ours)}
            & \textbf{25.11} & \textbf{37.83} & 8.29
            & \textbf{23.02} & \textbf{34.11} & \underline{7.87} \\
            \bottomrule
        \end{tabular}%
    }
\end{minipage}\hfill
\begin{minipage}[t]{0.44\textwidth}
\centering
\caption{Average per-step and per-episode time (seconds).}
\label{tab:efficiency}
\setlength{\tabcolsep}{4pt}
\resizebox{\linewidth}{!}{%
\begin{tabular}{@{}lcc@{}}
\toprule
Method & Step (s)\,$\downarrow$ & Episode (s)\,$\downarrow$ \\
\midrule
WMNav & 4.50 & 675.70 \\
APEX & 0.97 & 145.50 \\
PRPSearcher & 7.48 & 698.60 \\
OctMem-Agent & \underline{0.58} & \underline{62.30} \\
\midrule
\rowcolor{blue!8}
\textbf{\method{} (Ours)} & \textbf{0.31} & \textbf{34.50} \\
\bottomrule
\end{tabular}%
}
\end{minipage}
\end{table*}

\subsection{Three-Stage Training Curriculum}
\label{sec:training}

\paragraph{Stage 1: world-state transition.} Random-action rollouts train the BEV encoder, feature fusion, RSSM, and decoders, with the action head inactive and the pretrained visual and text encoders frozen. The RGB decoder reconstructs the observation, the BEV decoder reconstructs the exploration and obstacle layers, and the value decoder regresses onto the task-aware target of \eqref{eq:valuetarget} through the loss of \eqref{eq:valueloss}, while a Kullback--Leibler (KL) regularizer $\beta_{\mathrm{KL}}$ pulls the posterior toward the prior. Spatial reconstruction encourages the latent to retain the search state used by the later stages.

\paragraph{Stage 2: search and stop from experts.} Expert demonstrations pair instruction-conditioned search trajectories with terminal stopping decisions. Behavior cloning (BC) trains the action head to approach the target and issue \texttt{stop}, while the trainable world-model components continue to learn from reconstruction and value supervision. DINOv2 and SigLIP remain frozen. The resulting task policy $\pi_{\mathrm{BC}}$ initializes the actor refined in Stage~3.

\paragraph{Stage 3: value-guided imagination refinement.} With the RSSM and value decoder frozen, the world model rolls out prior-only trajectories from posterior states of real data via \eqref{eq:imagine}. Each candidate movement action $a$ yields an imagined successor state, from which the cognition branch decodes the future value layer $\hat V^{\mathrm{sp},a}_{t+1}$. Pooling this layer over the action's motion footprint $K^{a}_{t+1}$ on the BEV grid gives the action utility $q_V(s_t, a) = \sum_{x \in \mathcal{G}} K^{a}_{t+1}(x)\, \hat V^{\mathrm{sp},a}_{t+1}(x)$, discounted over $H=8$ steps with $\gamma=0.99$ to obtain $Q_V^{(H)}$. Superscript $a$ denotes action-conditioned successors. These utilities guide policy distillation without a critic loss.

The footprint $K^{a}_{t+1}$ is a binary mask of the sensor fan swept by the candidate motion, transformed from the UAV frame into the allocentric grid. Translation sweeps the fan along the motion, rotation sweeps it in place, and vertical movement projects it at the successor altitude. Out-of-grid cells are discarded. Thus, the readout combines a learned prediction of spatial utility with a deterministic geometric description of which region an action covers. Appendix~\ref{app:theory} gives the full construction.

The target policy reweights the behavior-cloned prior,
\begin{equation}
\label{eq:piplus}
\pi^{+}(a \mid s_t, e_\ell) \;\propto\; \pi_{\mathrm{BC}}(a \mid s_t, e_\ell)\, \exp\big(Q_V^{(H)}(s_t, a) / \tau\big),
\end{equation}
with temperature $\tau=1000$. The action branch matches this target by
\begin{equation}
\label{eq:actionloss}
\mathcal{L}_{\mathrm{action}} \;=\; D_{\mathrm{KL}}\big(\pi^{+}(\cdot \mid s_t, e_\ell)\, \big\|\, \pi_\theta(\cdot \mid s_t, e_\ell)\big),
\end{equation}
on movement actions only, with \texttt{stop} held at $\pi_{\mathrm{BC}}$ and the movement probabilities renormalized accordingly. Only the action head receives gradients.

\paragraph{Why retain the imitation prior?} The reweighting in \eqref{eq:piplus} solves a KL-regularized utility maximization,
\begin{equation}
\label{eq:utility_objective}
\max_{\pi}\; \mathbb{E}_{a\sim\pi}\!\left[Q_V^{(H)}(s_t,a)\right]
-\tau D_{\mathrm{KL}}\!\left(\pi\,\|\,\pi_{\mathrm{BC}}\right),
\end{equation}
with the stopping probability fixed at its imitated value. The KL term anchors movement refinement to demonstrated behavior while allowing higher predicted utility to shift probability among movement actions. A large $\tau$ recovers the imitation prior; a small $\tau$ concentrates movement probability on higher-utility actions. The derivation is provided in Appendix~\ref{app:theory}.

\paragraph{Deployment.} At each real step, depth and pose update the exploration and obstacle memory, the current observation updates the posterior belief $(h_t, z_t)$, and the final Stage-3 policy $\pi_\theta(\cdot \mid s_t, e_\ell)$ directly outputs the action distribution. No candidate rollouts, $H$-step utility evaluation, or value-based action ranking are performed at test time.

\section{Experiments}
\label{sec:experiments}
\begin{table*}[!t]
\caption{Ablations of architecture, spatial memory, and training stages on UAV-ON (\%). Bold and underlining mark the best and second-best values within each question group, respectively.}
\label{tab:ablation}
\centering
\setlength{\tabcolsep}{3pt}
\small
\begin{tabular*}{\textwidth}{@{\extracolsep{\fill}}lccc@{}}
\toprule
Configuration & SR$\uparrow$ & OSR$\uparrow$ & SPL$\uparrow$ \\
\midrule
\multicolumn{4}{@{}l}{\emph{(Q1) World-model architecture}} \\
\quad direct imitation (BC, no world model) & \underline{14.33} & \textbf{25.68} & \underline{7.20} \\
\quad X-Mobility-style (RSSM + imitation only) & 12.91 & 19.88 & 6.95 \\
\quad Dreamer-style (scalar critic in imagination) & \textbf{19.07} & \underline{24.99} & \textbf{7.64} \\
\midrule
 \multicolumn{4}{@{}l}{\emph{(Q2) Explicit spatial-cognition memory}} \\
\quad w/o explicit memory (latent-only belief) & 15.96 & 25.03 & 6.90 \\
\quad w/o exploration layer & \underline{18.09} & \underline{28.88} & \underline{7.54} \\
\quad w/o obstacle layer & \textbf{22.46} & \textbf{31.53} & \textbf{7.67} \\
\midrule
\multicolumn{4}{@{}l}{\emph{(Q3) Three-stage curriculum}} \\
\quad w/o Stage 1 (no random-action pretraining) & \underline{13.42} & 22.99 & 6.27 \\
\quad w/o Stage 2 (no expert imitation) & 10.97 & \underline{23.54} & \underline{6.80} \\
\quad w/o Stage 3 (no value-guided imagination) & \textbf{19.71} & \textbf{30.66} & \textbf{7.08} \\
\bottomrule
\end{tabular*}
\end{table*}

\begin{figure*}[!t]
\centering
\includegraphics[width=0.98\textwidth]{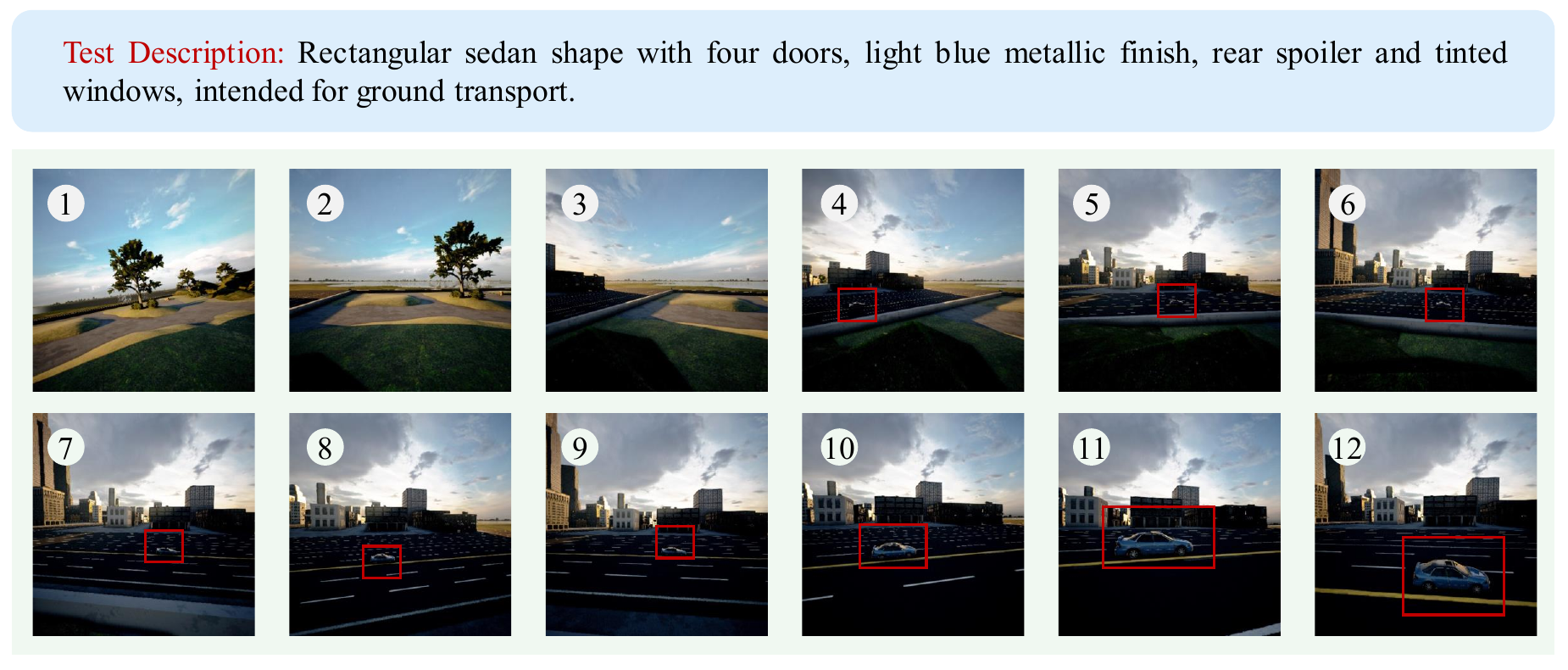}
\caption{A successful search for the instruction-specified light-blue sedan. Selected observations are shown in temporal order; red boxes mark the target.}
\label{fig:bev}
\end{figure*}

\subsection{Experimental Setup}
\label{sec:setup}

\paragraph{Benchmark and protocol.} UAV-ON \citep{xiao2025uavon} uses AirSim \citep{shah2018airsim} with 14 Unreal Engine scenes: 10{,}000 training episodes cover 10 scenes, and 1{,}000 test episodes cover all 14, including four held-out scenes. Each instruction specifies a target instance. \method{} uses eight fixed-step actions (Appendix~\ref{app:discretization}) and a 150-step limit. Success requires \texttt{stop} within $\tau_d=20\,\mathrm{m}$ of the target; stopping, collision, or budget exhaustion ends an episode.

\paragraph{Metrics.} Success Rate (SR) measures successful termination; Oracle Success Rate (OSR) measures entry into the target vicinity at any point; Success weighted by Path Length (SPL)~\citep{anderson2018spl} combines success and path efficiency. We report target-size and Seen/Unseen breakdowns; Appendix~\ref{app:metrics} defines the metrics.

\paragraph{Baselines.} We reproduce WMNav \citep{nie2025wmnav} and PRPSearcher \citep{ji2026cityavos} on UAV-ON, and use published results for Random, CLIP-H, AOA-F, AOA-V \citep{xiao2025uavon}, APEX \citep{zhang2026apex}, and OctMem-Agent \citep{zhou2026octmem}. Quoted results retain their original settings, including OctMem-Agent's 100-step limit. Appendix~\ref{app:metrics} details baseline provenance and controls; Appendix~\ref{app:implementation} gives our implementation.

\paragraph{Training data and implementation.} Stage~1 uses random trajectories of up to 50 steps. Stage~2 uses successful demonstrations from a rule-based expert combining target-bearing navigation and depth-based avoidance, stopping within $5\,\mathrm{m}$ (Appendix~\ref{app:data_collection}). Target coordinates serve only the training expert and value labels; policy inputs are front-view RGB-D, pose, and instruction. The RSSM has 1024-dimensional recurrent and 512-dimensional stochastic states. All stages use AdamW with frozen DINOv2 and SigLIP. Training takes approximately 20 hours on eight NVIDIA H20 GPUs; evaluation uses one NVIDIA RTX 5090.

\subsection{Main Results}
\label{sec:main_results}

\subsubsection{Search performance}
\label{sec:main_comparison}

\paragraph{Overall search performance.} \method{} achieves the highest overall SR ($23.80\%$) and OSR ($35.50\%$) in Table~\ref{tab:object_size_performance}, exceeding OctMem-Agent by $4.30$ and $6.20$ points. The joint gains are consistent with the intended roles of coverage memory and task-aware utility: retaining search progress and prioritizing useful future observations. SPL also increases from $6.37\%$ to $8.03\%$. APEX leads SPL ($10.14\%$), with a lower SR of $13.33\%$. Since SPL weights each success by path efficiency, \method{} completes more searches, while APEX scores higher after path-length weighting. Our utility rewards exploration and target proximity rather than directly minimizing route length, which may contribute to this difference in metric rankings.

\paragraph{Performance across target sizes.} \method{} leads SR and OSR in all three size groups, and SPL among methods reporting size-specific results. Its largest SR gain over OctMem-Agent is on large targets: $35.02\%$ versus $23.60\%$. Small- and medium-target SR reach $21.60\%$ and $21.80\%$, gaining $2.69$ and $3.71$ points. The OSR--SR gap is only $2.57$ points for large targets, versus $13.65$ for small targets. Thus, reaching the target vicinity translates into successful termination more consistently for large targets. More persistent visual evidence from larger objects is a plausible explanation for the stronger completion rate.

\paragraph{Generalization to unseen scenes.} \method{} leads SR and OSR on both splits in Table~\ref{tab:seen_unseen_comparison}. Held-out SR and OSR reach $23.02\%$ and $34.11\%$, exceeding OctMem-Agent by $5.45$ and $5.52$ points. Unseen SPL is $7.87\%$, versus $5.15\%$ for OctMem-Agent and $13.03\%$ for APEX. The seen-to-unseen decreases are $2.09$ SR points and $3.72$ OSR points, indicating that the policy retains both target discovery and successful termination under held-out layouts. Transfer is consistent with the memory design: coverage and obstacle layers are rebuilt from current depth and pose, retaining the same spatial meanings across layouts. The policy can therefore use a common representation of search progress in both familiar and held-out environments.

\subsubsection{Computational efficiency}
\label{sec:efficiency}

Table~\ref{tab:efficiency} compares per-step and per-episode time from our evaluation platform and the corresponding papers. \method{} records $0.31\,\mathrm{s}$ per step and $34.50\,\mathrm{s}$ per episode, the lowest listed times. Each real step updates BEV memory and the posterior belief before evaluating the refined action head. Imagined action evaluation is absorbed into policy learning, so deployment does not repeat the $H$-step candidate rollouts used in training. This separates the cost of learning prospective behavior from action selection, which requires no image generation or per-step LLM/VLA query.

\subsection{Ablation Study}
\label{sec:ablations}

Table~\ref{tab:ablation} examines \textbf{(Q1)} architecture, \textbf{(Q2)} spatial memory, and \textbf{(Q3)} training stages, using the same source datasets and action space. Evaluation excludes CityPark and BrushifyRoad because of invalid depth outputs documented by UAV-ON (Appendix~\ref{app:metrics}). Tables~\ref{tab:object_size_performance} and~\ref{tab:seen_unseen_comparison} report the complete system's benchmark results.

\paragraph{(Q1) World-model architecture.} Direct imitation reaches $14.33\%$ SR, the X-Mobility-style \citep{liu2024xmobility} RSSM with imitation reaches $12.91\%$, and the Dreamer-style \citep{hafner2020dreamer} variant reaches $19.07\%$ SR and $7.64\%$ SPL. The ordering suggests that latent dynamics become more useful when coupled to behavior optimization: imitation reproduces demonstrated actions, whereas imagined evaluation supplies signals for comparing action alternatives. Dreamer-style OSR ($24.99\%$) is slightly below direct imitation ($25.68\%$) despite its higher SR, locating the advantage in successful completion rather than increased target-vicinity coverage alone. These comparisons distinguish learning a predictive state from learning how to use it: the latter requires an objective over candidate actions. \method{} supplies this connection through utilities pooled from imagined spatial value layers.

\paragraph{(Q2) Explicit spatial memory.} The latent-only variant reaches $15.96\%$ SR. Retaining only obstacle memory reaches $18.09\%$, while retaining only exploration memory reaches $22.46\%$; both improve all three metrics over latent-only memory. The stronger exploration-only result is consistent with the information carried by coverage history: it distinguishes unexplored regions from previously observed free space, which occupancy alone does not encode. This distinction provides a direct cue for allocating search effort. Obstacle memory supplies a different cue, representing blocked geometry. The respective SR gains of $2.13$ and $6.50$ points suggest that retaining search progress is especially useful in this benchmark: free space identifies feasible motion, while coverage history helps distinguish where further exploration is worthwhile.

\paragraph{(Q3) Three-stage curriculum.} Omitting world-model pretraining yields $13.42\%$ SR; omitting expert imitation yields $10.97\%$, the lowest among stage-removal variants. Their OSRs are close ($22.99\%$ and $23.54\%$), but the no-imitation variant has lower SR, suggesting that imitation contributes to completing the task after reaching the target vicinity. Retaining both stages without imagination refinement yields $19.71\%$ SR and $30.66\%$ OSR. These results are consistent with distinct prerequisites for refinement: pretraining learns spatial transitions, while expert imitation establishes target-directed movement and stopping. In particular, Stage~3 refines movement while retaining the imitated stopping probability at a given belief, explaining why spatial-utility optimization relies on a task-policy foundation. It builds on these priors using the decoded utilities in Section~\ref{sec:training}.

\subsection{Case Study}
\label{sec:case_study}

Figure~\ref{fig:bev} follows \method{} in a successful search for the instruction-specified light-blue sedan. The UAV first surveys the roadside area, then approaches the target as it becomes visible across the road, and terminates with a successful stop. A representative failure and further analysis are provided in Appendix~\ref{app:visualization}.

\section{Conclusion}
\label{sec:conclusion}
We presented \method{}, a world model for aerial object search that connects explicit spatial memory with training-time imagination. Its decoded spatial utility guides movement-policy distillation while retaining an imitated stopping rule at each belief. On UAV-ON, the complete system achieves $23.80\%$ SR and $35.50\%$ OSR, with $23.02\%$ SR on held-out scenes. By transferring imagined utility into the action head during training, \method{} directly executes the learned policy at test time, with a measured latency of $0.31\,\mathrm{s}$ per step.\label{main-text-end}

{\small
\bibliographystyle{ieeenat_fullname}
\bibliography{references}
}
\clearpage
\appendix
\setcounter{figure}{0}
\setcounter{table}{0}
\renewcommand{\thefigure}{\Alph{section}.\arabic{figure}}
\renewcommand{\thetable}{\Alph{section}.\arabic{table}}

\section{Implementation Details}
\label{app:implementation}

\paragraph{Encoders.} The RGB front view is encoded twice: a frozen DINOv2 ViT-S/14 \citep{oquab2023dinov2} (cls token concatenated with the mean patch token, 768-d) provides spatial and geometric structure, and the vision tower of a frozen SigLIP2 \citep{zhai2023siglip} (768-d, L2-normalised) provides language-aligned features that share an embedding space with the instruction. Text: the target description is embedded offline by the same SigLIP2 text tower (768-d). A BEV encoder (five strided convolutions followed by a projection MLP, $2{\to}1024$-d) embeds the exploration and obstacle layers of the memory; this branch carries the episode-level history and the UAV's spatial situation. The streams ($768{+}768{+}768{+}1024$) are fused by a linear layer into the observation embedding $e_t$.

\paragraph{World model.} GRU hidden size 1024; stochastic state 512-d Gaussian (diagonal); posterior and prior heads are 2-layer multilayer perceptrons (MLPs) with softplus outputs for scale. Action embedding table: $8 \times 64$. Heads: an RGB decoder (StyleGAN-style generator with a perceptual + L1 loss), a BEV decoder that maps the latent state to the $256\times256$ exploration and obstacle layers (binary-layer losses), a value decoder that regenerates the task-aware spatial value layer (supervised by the target of \eqref{eq:valuetarget}), and an action head (categorical over the 8 atomic actions, conditioned on the latent state and the instruction).

\paragraph{System specifications.} Table~\ref{tab:specs} lists the spatial-memory grid, the geometric construction thresholds, and the training regularizer weights. The obstacle layer is built by pinhole back-projection of the depth images (Section~\ref{sec:bev_memory}).

\begin{table}[h]
\centering
\small
\setlength{\tabcolsep}{5pt}
\begin{tabular}{@{}lll@{}}
\toprule
Component & Parameter & Value \\
\midrule
\multirow{4}{*}{BEV grid} & grid size & $256 \times 256$ \\
 & cell resolution & $0.4\,\mathrm{m}$ \\
 & region coverage & $102.4\,\mathrm{m} \times 102.4\,\mathrm{m}$ \\
 & grid centering & episode start \\
\midrule
\multirow{2}{*}{Exploration layer} & sensing range $\rho$ & $20\,\mathrm{m}$ \\
 & FOV width $\phi$ & $90^{\circ}$ \\
\midrule
Obstacle layer & height threshold $z_{\mathrm{obs}}$ & $0.5\,\mathrm{m}$ \\
\midrule
Stage-1 training & KL weight $\beta_{\mathrm{KL}}$ & $10^{-3}$ \\
\bottomrule
\end{tabular}
\caption{System specifications of \method{} referenced in Section~\ref{sec:bev_memory} and Section~\ref{sec:training}.}
\label{tab:specs}
\end{table}

\paragraph{Value supervision.} The target of \eqref{eq:valuetarget} uses $\lambda_{\mathrm{expl}}=\lambda_{\mathrm{goal}}=1.0$ and a goal-centered Gaussian with $\sigma_g=10\,\mathrm{m}$.

\paragraph{Reference implementation and configuration.} The following settings describe the supplied reference implementation. Stages~1 and~2 are configured for 100 and 30 epochs, respectively, with per-device batch sizes of 8 and 4; Stage~2 accumulates gradients over three batches and weights action classification by 10. The pretrained image and text backbones remain frozen. All stages use AdamW \citep{loshchilov2019adamw} with weight decay $0.01$. Batch sizes are per device; the global batch additionally scales with the number of training devices. The Stage-3 configuration specifies 30 epochs, batch size 8, sequence length 8, an initial learning rate of $10^{-4}$, AdamW weight decay $0.01$, and action-loss weight $1.0$. A cosine schedule is stepped after each optimizer update, and the global gradient norm is clipped at $1.0$. No gradient accumulation is configured for Stage~3. Stages~1 and~2 use a per-update one-cycle schedule with peak learning rate $10^{-5}$ and the increasing phase occupying $20\%$ of updates. Mixed precision uses FP16 autocasting. Stage~3 selects checkpoints by validation action loss and retains the two best checkpoints and the last checkpoint. In the accompanying implementation, the KL regularizer gives equal weight to the posterior and prior gradient terms.

\paragraph{Network dimensions and shared spatial decoding.} The frozen image backbones are DINOv2-S/14 and SigLIP2-Base/16 with a $224\times224$ SigLIP2 input. The observation-fusion projection maps $3328$ features to $1024$ dimensions. The BEV encoder uses $4\times4$, stride-2 convolutions with channel widths $(32,64,128,256,512)$, each followed by batch normalization and ReLU, then an MLP maps the flattened $512\times8\times8$ feature to $2048$ and then $1024$ dimensions. The deterministic and stochastic states concatenate to $1536$ dimensions. Gaussian scales use $\operatorname{softplus}(u)+0.1$. In addition to the action input to the GRU, the prior and posterior MLPs receive separate learned projections of the 64-dimensional action embedding. The spatial and value outputs share one instruction-conditioned BEV decoder: the concatenation of the latent state and the 768-dimensional instruction is projected to $512\times8\times8$ features, then upsampled through widths $(256,128,64,32)$ to three $256\times256$ output channels. Sigmoid activations bound all three outputs to $[0,1]$. Auxiliary $1\times1$ heads provide predictions at resolutions $16$, $32$, $64$, and $128$. The action head projects latent state and instruction into two 2048-dimensional tokens, fuses them with four-head self-attention, and applies an MLP with hidden widths $(2048,2048,1024)$ to produce eight logits; the auxiliary path output is not used by the discrete-action objective.

\paragraph{Spatial reconstruction losses.} The exploration and obstacle channels use class-weighted binary cross-entropy plus Dice loss with coefficient $1.0$; frames with an entirely zero target for a channel are skipped for that channel. The positive-class weight is the negative-to-positive pixel ratio, clipped to $[0.1,10]$. The value channel uses smooth-L1 loss with transition parameter $0.1$. Losses at resolutions $(256,128,64,32,16)$ have weights $(1,1/2,1/4,1/8,1/16)$, respectively. The three channel losses have equal internal weights. The configured outer BEV loss weight is $1.0$ in Stage~1 and $3.0$ in Stage~2; the RGB reconstruction weight is $1.5$ in both configurations. Stage~3 uses only the action-refinement objective and does not optimize reconstruction losses. Stage~1 also configures a six-class auxiliary segmentation loss with weight $2.0$. The converter reads front-view label maps from \texttt{semantic\_front.npy}.

\paragraph{Sequence handling and deployment.} Training uses non-overlapping eight-frame windows within each parquet file, discarding incomplete trailing windows. The RSSM initializes its recurrent and stochastic states to zero at each window and detaches them between successive observed steps. The first observation uses action ID 0 as a fixed preceding-action placeholder; subsequent observations use the recorded previous action. Stage~3 encodes the real sequence with posterior means and starts imagination from its final posterior state and recorded planar pose. Imagined transitions sample the Gaussian prior. The frozen world model stays in evaluation mode, and the refined actor is initialized as a trainable copy of the Stage-2 action head. Deployment uses the streaming observation-update interface, which accepts and returns the recurrent state; its posterior uses the Gaussian mean. Action selection is direct and does not invoke the imagination engine. The refined actor must be included in the deployment checkpoint rather than leaving the original frozen Stage-2 action head in place.

\paragraph{Planar footprint rasterization.} Each motion sweep is approximated by the union of five equally spaced sensor-fan samples, including both endpoints, before value pooling. Utilities are sums over in-grid cells rather than area-normalized means. The footprint state is $(x,y,\psi)$: ascent and descent leave the planar footprint unchanged, while the RSSM receives their distinct action IDs and can predict different spatial value layers. The fixed allocentric grid is centered at the episode start, with column $u=128+X/0.4$ and row $v=128-Y/0.4$ for coordinates in meters relative to that origin. The data converters rotate both the UAV and target displacements by the negative initial yaw, so $+X$ follows the starting heading. For example, at an initial yaw of $90^\circ$, a $5$\,m displacement along world $+Y$ maps to $(5,0)$\,m. Depth back-projection uses yaw and a fixed camera-height proxy of $1$\,m. A pure climb changes neither the planar pose nor this height proxy.

\section{Action Discretization Protocol}
\label{app:discretization}

We discretize UAV-ON's parameterized control interface into eight fixed-step actions: five translations (\texttt{forward}/\texttt{left}/\texttt{right} = $d_{\mathrm{trans}}$ meters, \texttt{ascend}/\texttt{descend} = $d_{\mathrm{vert}}$ meters), two rotations (\texttt{rotate-left}/\texttt{rotate-right} = $\theta_{\mathrm{rot}}$ degrees), and \texttt{stop}, which terminates the episode and triggers the success evaluation (success iff the UAV is within threshold distance $\tau_d = 20\,\mathrm{m}$ of the target at stopping time, per the official UAV-ON criterion), with $d_{\mathrm{trans}} = 5\,\mathrm{m}$, $d_{\mathrm{vert}} = 2\,\mathrm{m}$, $\theta_{\mathrm{rot}} = 15^{\circ}$. These magnitudes define our fixed-step implementation, not a benchmark-mandated discretization (Section~\ref{sec:formulation}). This discrete interface yields a compact, embeddable action space for the world model: each action maps to a single learned embedding (Section~\ref{sec:world_model}), and its fixed motion delta makes the footprint of any candidate action computable in closed form (Appendix~\ref{app:theory}).

\section{Training-Time Imagination and Footprint Geometry}
\label{app:theory}

This section specifies the implemented policy targets and geometric constructions introduced in Sections~\ref{sec:value_imagination} and~\ref{sec:training}.

\paragraph{Movement-policy targets.} At each imagined state, the implementation forms a target conditional distribution over the seven movement actions by combining the frozen imitation prior with decoded spatial utilities. The training loss compares this target with the actor's conditional movement distribution. The full target retains the imitation prior's stopping mass; this target construction does not itself impose an equality constraint on the refined actor's stopping probability. At deployment the refined actor emits all eight logits, so its stopping probability can differ from the imitation prior at the same belief.

\paragraph{Imagined training states.} An actor-sampled trajectory supplies $H=8$ imagined training states. At each state, seven one-step candidate transitions produce action-conditioned spatial utilities. The implementation adds the same discounted continuation term to every candidate at that state. This common term cancels in the normalized reweighting, so the candidate-dependent supervision is the one-step spatial utility. The horizon controls the extent of imagined-state training rather than a separate full-depth branch for each candidate. Holding those states and the one-step samples fixed, changing the continuation discount leaves the movement targets unchanged.

\paragraph{Action footprint geometry.} The Stage-3 utility pooling operator of Section~\ref{sec:value_imagination} aggregates the decoded value over a per-action \emph{motion footprint} on the same allocentric grid as the memory. Formally, let $\omega : \mathbb{R}^2 \to \mathcal{G}$ be the fixed affine map from world coordinates to grid cells (Section~\ref{sec:bev_memory}), and let each atomic action $a$ carry its implementation-fixed motion delta $\Delta_a$ (Appendix~\ref{app:discretization}: $5\,\mathrm{m}$ horizontal translations, $2\,\mathrm{m}$ vertical translations, and $15^{\circ}$ rotations). The canonical footprint $F^{a} \subset \mathbb{R}^2$ is defined in the ego frame as the sensor fan of half-width $\phi/2$ and range $\rho$ (Section~\ref{sec:bev_memory}) swept along $\Delta_a$; for rotation actions the fan rotates in place by $\theta_{\mathrm{rot}}$, and for vertical actions the two-dimensional footprint is unchanged. Given the planar position $(x_t,y_t)$ and yaw $\psi_t$ extracted from the known pose $p_t$, the allocentric footprint is $K^{a}_{t+1}(x) = \mathbb{1}\!\left[\, \omega^{-1}(x) \in R(\psi_t) F^{a} + (x_t, y_t) \,\right]$, where $R(\psi_t)$ is the planar rotation by the yaw and the successor pose follows from composing $p_t$ with $\Delta_a$; cells falling outside the grid are dropped, so the pooled sum runs over in-grid cells only. The action utility $q_V(s_t, a) = \sum_{x} K^{a}_{t+1}(x)\,\hat V^{\mathrm{sp},a}_{t+1}(x)$ is therefore the expected value mass that action $a$ sweeps under the sensing geometry, in the same world frame in which the value layer is decoded. Because every $\Delta_a$ is fixed in our implementation, the footprint of each candidate action is computable in closed form from the current pose; no component of it is learned.

\paragraph{Training imagination and direct deployment.} The horizon $H=8$ bounds the prior-only rollouts used to construct policy-improvement targets during Stage~3 training. At evaluation, each incoming observation updates the posterior belief $(h_t, z_t)$, and the final refined actor $\pi_\theta$ directly outputs the action distribution from this belief and the instruction. The recurrent state is updated at every real step over the full episode. Evaluation performs no candidate expansion, prior-only imagined rollout, or value-based action ranking; $H$ is a training horizon, not a test-time planning budget.

\section{Evaluation Metrics}
\label{app:metrics}

Let $\mathcal{E} = \{e_i\}_{i=1}^{N}$ denote the evaluation set of $N$ episodes. For episode $e_i$, let $S_i \in \{0,1\}$ indicate success as defined in Section~\ref{sec:setup} (a \texttt{stop} issued within $\tau_d$ of the target), $O_i \in \{0,1\}$ indicate oracle success (the agent's trajectory passed within $\tau_d$ of the target at any step, regardless of where it stopped), $p_i$ the length of the executed path (cumulative movement distance), $l_i$ the benchmark-provided geodesic distance from the start to the target.

\paragraph{Success Rate.}
\begin{equation}
\mathrm{SR} = \frac{1}{N} \sum_{i=1}^{N} S_i .
\end{equation}

\paragraph{Oracle Success Rate.}
\begin{equation}
\mathrm{OSR} = \frac{1}{N} \sum_{i=1}^{N} O_i .
\end{equation}
The gap $\mathrm{OSR} - \mathrm{SR}$ measures episodes that enter the target vicinity but do not terminate successfully there. It does not by itself distinguish recognition errors from later departure, collision, or failure to stop.

\paragraph{Success weighted by Path Length} \citep{anderson2018spl}:
\begin{equation}
\mathrm{SPL} = \frac{1}{N} \sum_{i=1}^{N} S_i \,
\frac{l_i}{\max(p_i, \, l_i)} .
\end{equation}
SPL credits only successful episodes and penalizes path lengths that exceed the geodesic optimum; a value of $1$ would mean every episode succeeded along a shortest path.

\paragraph{Offline diagnostic scripts.} The supplied offline replay tools operate on recorded observations and do not replace closed-loop UAV-ON evaluation. Their default path-efficiency calculation uses planar straight-line reference distance when no geodesic distance is supplied. The SR, OSR, and SPL definitions above specify the benchmark quantities; offline replay outputs are diagnostic approximations. The timing utility measures a model step on synthetic, preassembled inputs. The utility measures model inference only. Episode wall-clock measurements require closed-loop execution.

\paragraph{Ablation evaluation protocol.} For the ablated variants in Table~\ref{tab:ablation}, we exclude evaluation episodes from CityPark and BrushifyRoad because their depth outputs do not reflect scene geometry, as documented in the official UAV-ON repository's ``UAV-ON 1.0 Environment Note.'' Table~\ref{tab:ablation} compares variants on this subset. Subtracting a subset score from the full-set score in Table~\ref{tab:object_size_performance} mixes the change in method with the change in episodes.

\paragraph{Baseline provenance.} APEX results in Table~\ref{tab:object_size_performance} and Table~\ref{tab:seen_unseen_comparison} are taken from \citet{zhang2026apex}. Its paper reports overall, seen, and unseen metrics only, without a per-object-size breakdown, which is why the per-size cells for APEX in Table~\ref{tab:object_size_performance} are left blank. The UAV-ON benchmark baselines (AOA-F, AOA-V, CLIP-H) are taken directly from the benchmark paper \citep{xiao2025uavon}, and OctMem-Agent \citep{zhou2026octmem} likewise from its paper, retaining its reported six-action configuration and 100-step episode limit. APEX's overall and split scores retain the aggregation reported in its original paper and are not recomputed using our test partition. WMNav and PRPSearcher are reproduced by us on UAV-ON; the multimodal language models invoked by the reproduced baselines use Qwen3.6-Plus. The quantitative study focuses on UAV-ON agents and the reproduced navigation baselines listed in the tables; video-generation approaches are discussed in Section~\ref{sec:related}. SearchWorld uses eight actions and a 150-step limit, so its overall SR difference from OctMem-Agent includes differences in control and search budget.

\begin{figure*}[t]
\centering
\includegraphics[width=0.98\textwidth]{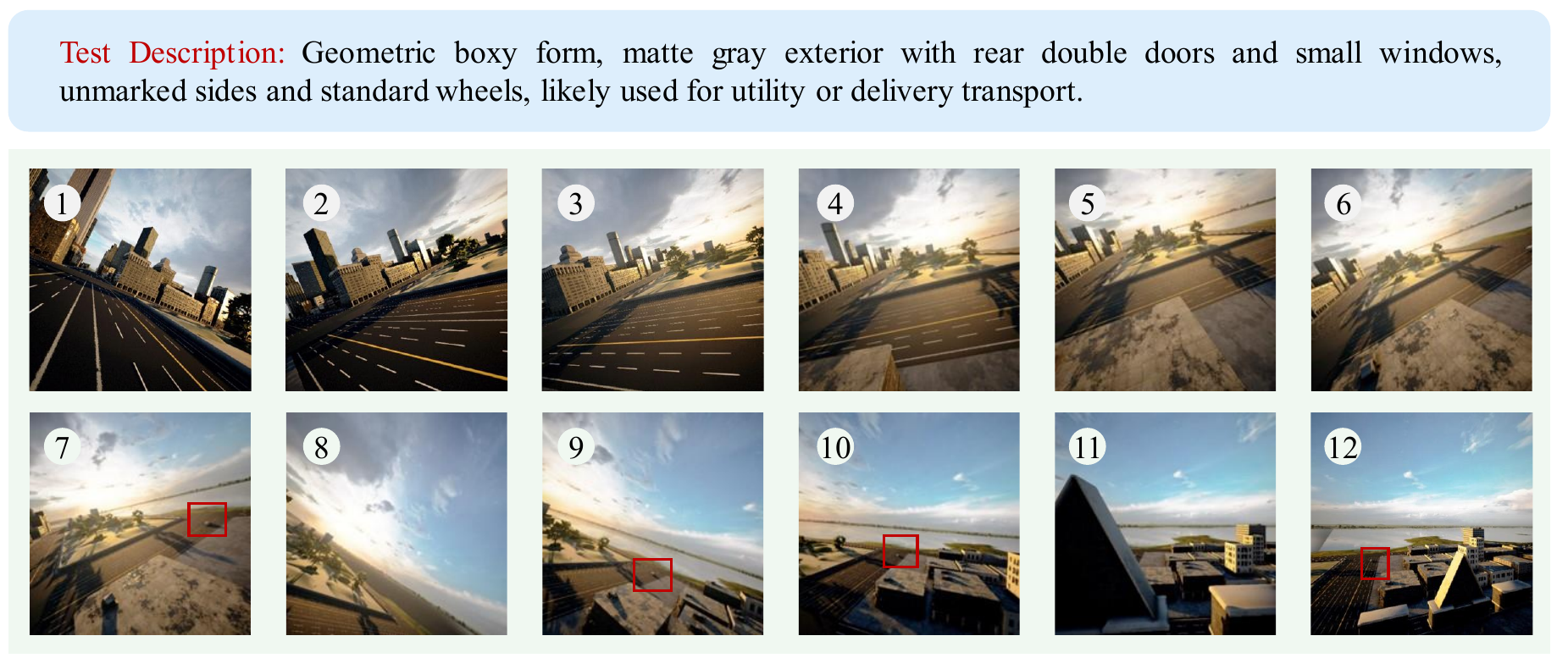}
\caption{A representative failure case. The instruction names the target instance (a gray utility box truck). The UAV explores its surroundings in the early panels but stays too far from the target to detect it, and the step budget is exhausted in the complex environment.}
\label{fig:failure}
\end{figure*}

\section{Training Data Collection}
\label{app:data_collection}

Both datasets used in the three-stage curriculum (Section~\ref{sec:training}) were collected in the UAV-ON AirSim environments \citep{shah2018airsim}. Collection records 4-view RGB-D ($512{\times}512$), 6-DoF pose, IMU readings, and the discretized atomic action set. The spatial memory is constructed from the front-view depth (Section~\ref{sec:bev_memory}); the remaining views are recorded but not consumed by the current architecture. All episodes record per-step position, attitude (quaternion and Euler), linear and angular velocities, distance to target, cumulative movement, collision flags, and oracle-success flags, so that transitions are fully supervised.

\paragraph{Random-interaction dataset (Stage 1).} For world-model pretraining we collect random rollouts: at each step an action is drawn uniformly from the seven movement primitives (an optional forward-biased distribution and a low-probability \texttt{stop} are supported but disabled by default), for up to $50$ steps per episode. Episodes that collide are discarded and re-collected from the same initial state (up to 10 attempts) so that the dataset contains only physically valid trajectories; the surviving episodes carry the full sensor stream, including float32 depth in meters. Each episode is stored with a summary (outcome, final distance, action histogram), and collection supports per-scene checkpointing so runs can be resumed after simulator restarts.

\paragraph{Expert-demonstration dataset (Stage 2).} Behavior cloning requires successful demonstrations. Since UAV-ON ships annotations but no trajectories, we generate them with a rule-based expert that combines oracle bearing navigation with reactive depth-based avoidance. The expert is a five-phase priority machine: (1)~\texttt{STOP} when the 3-D distance to the target falls below $r_{\mathrm{stop}} = 5\,\mathrm{m}$, tighter than the evaluation threshold $\tau_d = 20\,\mathrm{m}$, so that demonstrations end well inside the success region rather than at its boundary; (2)~\texttt{ALIGN}: rotate toward the target whenever the bearing error exceeds $7.5^{\circ}$ (set to half the $15^{\circ}$ rotation increment); (3)~\texttt{HEIGHT\_ALIGN}: ascend or descend when horizontally close (within $25\,\mathrm{m}$) and the vertical offset exceeds a $3\,\mathrm{m}$ dead-band, with ground protection from the down-facing camera ($4\,\mathrm{m}$ minimum clearance) and an altitude ceiling of $60\,\mathrm{m}$; (4)~\texttt{APPROACH}: fly forward when frontal clearance exceeds $10\,\mathrm{m}$; (5)~\texttt{AVOID}: otherwise side-step toward the clearer lateral direction (threshold $8\,\mathrm{m}$), climb when both sides are blocked, with an anti-oscillation counter that forces a climb after repeated left--right flips. Frontal and lateral clearances are computed robustly as the 5th percentile of valid depth pixels in the central ROI of the corresponding camera.

Only episodes that end in a successful stop are kept; failed attempts are retried from the same initial state with $\epsilon$-safe exploration noise ($15\%$ probability of replacing the expert action with a random collision-safe one on retry attempts), which lets the otherwise deterministic expert escape repeated failures. Tasks that exhaust their retry budget are skipped and logged. A watchdog monitors the simulator (UE instances degrade after hours of uptime, e.g.\ constant-depth frames), restarts it, and resumes collection; spawn-mismatched tasks (where the engine relocates the UAV at reset) and underground spawns are detected and excluded automatically. The result is a corpus of successful demonstrations, one per covered task up to retry-budget exhaustion, stored in a compact per-episode format (JSONL state-action records, JPEG frames, and quantized depth stacks).

\section{Qualitative Visualizations and Failure Analysis}
\label{app:visualization}

The case-study of Figure~\ref{fig:bev} (Section~\ref{sec:case_study}) illustrates the successful search behavior discussed in the main text.

\paragraph{A representative failure case.} Figure~\ref{fig:failure} traces a representative failure. Early in the episode the UAV explores its surroundings, but it stays too far from the target to detect it; the refined policy continues exploring sectors that do not contain the target until the step budget is exhausted in the complex environment. This case illustrates a remaining difficulty in directing exploration toward an initially unobserved target.

\paragraph{Failure taxonomy.} Qualitative inspection of test episodes identifies three recurring failure patterns of \method{}:
\begin{itemize}
\item \textbf{Occlusion traps}: the agent repeatedly approaches a region behind tall structures
without circumnavigating, exhausting the step budget. The obstacle
layer marks the blocking cells, but the learned movement policy fails
to route around the obstruction.
\item \textbf{Semantic confusion}: the agent stops near a
distractor object of the same category but wrong instance, suggesting difficulty in distinguishing
instances at distance.
\item \textbf{Frontier dithering}: the learned policy oscillates between
two unexplored directions instead of committing to sustained exploration.
\end{itemize}
Each mode suggests a targeted fix (occlusion-aware value supervision, instance-aware goal embeddings, and temporal consistency in the learned movement policy, respectively), which we leave to future work.

\section{Limitations}
\label{app:limitations}
\paragraph{(1) Simulation and sensing assumptions.} All experiments use the UAV-ON simulation environment. \method{} assumes exact 6-DoF pose, ground-truth depth, and fixed camera intrinsics supplied by the simulator. Errors in pose or depth can propagate into the deterministic memory layers of Section~\ref{sec:bev_memory}. The visual and language encoders (DINOv2 and SigLIP2) are frozen pretrained models. Robustness to sensor noise, pose perturbations, and transfer to real scenes or other simulators remains to be evaluated.

\paragraph{(2) Action execution.} The discrete action interface assumes idealized execution. Physical actuation dynamics, control latency, and execution noise are not modeled in the current evaluation. Deployment would require testing the policy with these effects and the corresponding flight controller.

\paragraph{(3) Obstacle representation and avoidance.} The two-dimensional BEV memory uses a height threshold and abstracts away fine-grained vertical structure, including overhead wires. The obstacle layer provides geometric context for the learned movement policy but does not guarantee collision-free motion. Handling such structures requires richer geometry and further evaluation of avoidance behavior.

\paragraph{(4) Value supervision and stopping.} The value decoder is conditioned on the latent state and instruction. Ground-truth target positions are used only to construct training labels, so the quality of this supervision affects movement-policy refinement. The deployed actor selects actions directly. The refinement loss acts on conditional movement probabilities; it does not explicitly constrain the actor's stopping probability. Movement refinement therefore relies on both the learned spatial utilities and the task behavior established through imitation.

\section{Ethics Statement}
This work studies autonomous UAV navigation for object search in simulation.
The study focuses on searching for specified objects and does not develop
person-tracking capabilities. Real-world use would require addressing
privacy, surveillance misuse, and flight safety. The compute used is reported in Section~\ref{sec:setup}.

\section{Reproducibility Statement}
All models are trained and evaluated on the public UAV-ON benchmark with
frozen public encoders (DINOv2, SigLIP). Implementation details, the
discretization protocol, and qualitative analyses are provided in
Appendices~\ref{app:implementation}--\ref{app:visualization}. Code will be made publicly available.

\end{document}